\documentclass{colt2013} 

\makeatletter
\AtBeginDocument{\Hy@breaklinkstrue}
\makeatother

\title[]{Stochastic Optimization of Smooth Loss}

   \coltauthor{\Name{Rong Jin} \Email{rongjin@cse.msu.edu}\\
    \addr Department of Computer Science and Engineering \\Michigan State University\\East Lansing, MI 48824, USA}

\usepackage{algorithm,algorithmic}

\newtheorem{thm}{Theorem}

\def \R {\mathbb{R}}
\def \x {\mathbf{x}}
\def \E {\mathrm{E}}

\def \F {\mathcal{F}}

\def \w {\mathbf{w}}
\def \wh {\widehat{\w}}

\def \D {\mathcal{D}}

\def \Dh {\widehat{\D}}

\def \lh {\widehat{\ell}}

\begin{document}

\maketitle

Let $\phi(z)$ be a smooth loss function, with $|\ell'(z)| \leq L$ and $|\ell'(z) - \ell'(z')| \leq \gamma |z - z'|$. Let $\Omega = \left\{\w \in \R^d: |\w| \leq R\right\}$ be the solution domain. Let $(\x_i, y_i), i=1, \ldots, n$ be the sequence of i.i.d samples used for training, where $\x_i \in \R^d$ and $y_i \in \{-1, +1\}$. Our goal is to find a solution $\wh$ with a good generalization performance. More specifically, let $\ell(\w)$ be the expected loss for any solution $\w$, i.e. $\ell(\w) = \E[\ell(y\w^{\top}\x)]$. Our goal is to minimize $\ell(\w)$.

A straightforward approach is to optimize $\ell(\w)$ by stochastic optimization. Let $w_1 = 0$ be the initial. At each iteration $t$, we receive a training example $(\x_i, y_i)$, and update the current solution $\w_t$ by
\[
\w_{t+1} = \mathop{\arg\min}\pi_{\Omega}\left(\w_t - \eta \nabla \ell_t(\w_t) \right)
\]
where $\eta > 0$ is the stepsize and $\ell_t(\w) = \phi(y_t\w^{\top}\x_t)$. The final solution $\wh$ will be the average of all the solutions, i.e. $\wh = \sum_{t=1}^T \w_t/T$. In~\citep{srebro:2010:smoothness}, the authors were able to show that a simple stochastic optimization method, with an appropriate choice of step size $\eta$, can achieves the following generalization error bound in expectation, i.e.
\[
\E[\ell(\wh)] \leq \ell(\w_*) + K\left(\frac{t}{n} + \sqrt{\ell(\w_*)\frac{t}{n}} \right)
\]
where $t = \gamma |\w_*|^2$.

There are two limitations with the analysis in~\citep{srebro:2010:smoothness}. First, it shows a bound in expectation, not a high probability bound. Second, it requires the knowledge of $\ell(\w_*)$ for tuning the step size in order to achieve the desired bound. In the draft presented in this work, we improve the analysis in~\citep{srebro:2010:smoothness} by addressing these two limitations.

First, let's address the first limitation by showing a high probability bound. At each iteration, we have
\begin{eqnarray*}
\ell_t(\w_t) - \ell_t(\w_*) & \leq & \frac{|\w_t - \w_*|^2}{2\eta} - \frac{|\w_{t+1} - \w_*|^2}{2\eta} + \frac{\eta}{2}|\nabla \ell_t(\w_t)|^2 \\
& \leq & \frac{|\w_t - \w_*|^2}{2\eta} - \frac{|\w_{t+1} - \w_*|^2}{2\eta} + 2\eta\gamma \ell_t(\w_t)
\end{eqnarray*}
where in the last step, we use the property $|\phi'(y_t\w_t^{\top}\x_t)|^2 \leq 4\gamma\phi(y_t\w_t^{\top}\x_t)$. By adding the inequalities of all iterations and using the assumption $\eta \leq 1/[2\gamma]$, we have
\begin{eqnarray*}
\lefteqn{\sum_{t=1}^T\ell(\w_t) - \ell(\w_*)) \leq} \\
& & \frac{R^2}{2\eta} + 2\eta\gamma \sum_{t=1}^T \ell(\w_t) + (-2\eta\gamma + 1)\underbrace{\sum_{t=1}^T \ell(\w_t) - \ell_t(\w_t)}_{:= A_T} + \underbrace{\sum_{t=1}^T \ell_t(\w_*) - \ell_t(\w_*)}_{:= B_T}
\end{eqnarray*}

To bound $A_T$ and $B_T$, we need the following bound for martingales.

\begin{thm} \label{thm:bernstein} (Bernstein¡¯s inequality for martingales). Let $X_1, \ldots , X_n$ be a bounded martingale difference sequence with respect to the filtration $\F = (\F_i)_{1\leq i\leq n}$ and with $\|X_i\| \leq K$. Let
\[
S_i = \sum_{j=1}^i X_j
\]
be the associated martingale. Denote the sum of the conditional variances by
\[
    \Sigma_n^2 = \sum_{t=1}^n \E\left[X_t^2|\F_{t-1}\right]
\]
Then for all constants $t$, $\nu > 0$,
\[
\Pr\left[ \max\limits_{i=1, \ldots, n} S_i > t \mbox{ and } \Sigma_n^2 \leq \nu \right] \leq \exp\left(-\frac{t^2}{2(\nu + Kt/3)} \right)
\]
and therefore,
\[
    \Pr\left[ \max\limits_{i=1,\ldots, n} S_i > \sqrt{2\nu t} + \frac{\sqrt{2}}{3}Kt \mbox{ and } \Sigma_n^2 \leq \nu \right] \leq e^{-t}
\]
\end{thm}

Using the above theorem, with a probability $1 - e^{-t}$, we can bound $B_T$ by
\[
B_T \leq \frac{\sqrt{2}t}{3}C + \sqrt{2tC\ell(\w_*) T}
\]
where $C = LR+\phi(0)$ and $t = \log(1/\delta)$. To bound $A_T$, we define martingale difference $X_t = \ell(\w_t) - \ell_t(\w_t)$. Define the conditional variance $\Sigma_T^2$ as
\[
    \Sigma_T^2 = \sum_{t=1}^{T} \E_{t}\left[X_t^2 \right] \leq \sum_{t=1}^{T} \E_t[\ell^2_t(\w_t)] \leq C\sum_{t=1}^T \ell(\w_t) = CD_T
\]
where $D_T := \sum_{t=1}^T \ell(\w_t)$. Using the Berstein inequality for martingale sum, we have
\begin{eqnarray*}
\lefteqn{\Pr\left(A_T\geq 2\sqrt{CD_T \tau} + \sqrt{2}C\tau/3\right)} \\
& = & \Pr\left(A_T\geq 2\sqrt{CD_T \tau} + \sqrt{2}C\tau/3, \Sigma_T^2 \leq CD_T\right) \\
& = & \Pr\left(A_T\geq 2\sqrt{CD_T\tau} + \sqrt{2}C\tau/3, \Sigma_T^2 \leq CD_T, D_T \leq C\right) \\
&  & + \sum_{i=1}^m \Pr\left(A_T \geq 2\sqrt{C D_T \tau} + \sqrt{2}C\tau/3, \Sigma_T^2 \leq CD_T, 2^{i-1}C < D_T  \leq 2^{i} C \right) \\
& \leq & \Pr\left(D_T \leq C \right) + \sum_{i=1}^m \Pr\left(A_T \geq C\sqrt{2^{i+1}\tau} + \sqrt{2}C\tau/3, \Sigma_T^2 \leq C2^i\right) \\
& \leq & \Pr\left(D_T \leq C\right) + me^{-\tau}
\end{eqnarray*}
where $m = \lceil \log_2 T \rceil$. As a result, we have
\[
\Pr\left(A_T \leq 2\sqrt{CD_T t} + \frac{\sqrt{2}}{3}Ct \right) + \Pr(D_T \leq C) \geq 1 - e^{-t}
\]
where $t = \log(1/\delta) + \log m$.

Using the bounds for $A_T$ and $B_T$, we have, with a probability $1 - 2e^{-t}$,
\begin{eqnarray*}
\lefteqn{(1 - 2\eta\gamma)D_T - \ell(\w_*) T \leq} \\
& & C + \frac{R^2}{2\eta} + (1 - 2\eta\gamma)\left(2\sqrt{CD_Tt} + \frac{\sqrt{2}}{3}Ct \right) + \frac{\sqrt{2}}{3}Ct + \sqrt{2tC\ell(\w_*)T}
\end{eqnarray*}
where $t = \log(1/\delta) + \log m$. Reorganizing the terms in the above inequality, we have
\[
(1 - 2\eta\gamma)\left(D_T - 2\sqrt{CD_Tt}\right) - \ell(\w_*)T \leq \frac{R^2}{2\eta} + Ct + \sqrt{2tC\ell(\w_*)T}
\]
where $t = \log(1/\delta) + \log m + 1$. It is easy to verify that $D_T - 2\sqrt{CD_T t}$ is monotonically increasing when $D_T \geq Ct$. Hence, we have, with a probability $1 - 2\delta$,
\[
(1 - 2\eta\gamma)\left(\ell(\wh) - 2\sqrt{\frac{Ct}{T} \ell(\wh)}\right) \leq \ell(\w_*) + \frac{R^2}{2\eta T} + \frac{Ct}{T} + \sqrt{\frac{2Ct}{T}\ell(\w_*)}
\]
or
\begin{eqnarray}
\ell(\wh) - \ell(\w_*) \leq \frac{R^2}{2\eta T} + 2\eta\gamma \ell(\wh) + \sqrt{\frac{2Ct}{T}\ell(\w_*)} + 2\sqrt{\frac{Ct}{T}\ell(\wh)} + \frac{Ct}{T} \label{eqn:bound-1}
\end{eqnarray}
By setting $\eta = R/2\sqrt{\gamma T \ell(\wh)}$, we have, with a probability $1 - \delta$,
\[
\ell(\wh) - \ell(\w_*) \leq \sqrt{\frac{2Ct}{T}\ell(\w_*)} + 2\sqrt{\frac{Ct}{T}\ell(\wh)} + \frac{Ct}{T}
\]
where $t = \log(1/\delta) + \log m + 1 + R^2\gamma/C$. Since $\ell(\w_*) \leq C$ and $\ell(\wh) \leq C$, under the assumption $T \geq t$, we have
\[
\ell(\wh) - \ell(\w_*) \leq \frac{Ct}{T} + 4C\sqrt{\frac{t}{T}} \leq 5C\sqrt{\frac{t}{T}}
\]
and therefore, with a probability $1 - 2\delta$
\[
\ell(\wh) - \ell(\w_*) \leq 4\sqrt{\frac{Ct}{T}\ell(\w_*)} + (2\sqrt{5} + 1)\frac{Ct}{T}
\]

The above analysis allows us to derive a high probability bound for the proposed algorithm. It however does not resolve the problem of determining the appropriate step size $\eta$. We address this limitation by exploring the doubling trick. We divide the learning process into $m$ epoches where the $k$th epoch is comprised of $T_k$ training examples, with $T_k = T_1 2^{k-1}$. Let $\w^1_k, \ldots, \w^{T_k}_k$ be the sequence of solutions generated by the $k$th epoch. Define
\[
D_k = \frac{1}{T_k}\sum_{i=1}^{T_k} \ell(w_k^i)
\]
We assume that, with a probability $1 - \delta$, we have
\[
D_k - \ell(\w_*) \leq K\left(\frac{Ct}{T_k} + \sqrt{\frac{Ct}{T_k}\ell(\w_*)}\right)
\]
where $t = \log(1/\delta) + \log m + 1 + R^2\gamma/C$. Define $\Dh_k$ as
\[
\Dh_k = \frac{1}{T_k} \sum_{i=1}^{T_k} \ell_k^i(\w_k^i)
\]
where $\ell_k^i(\w) = \phi(y_k^i\w^{\top}\x_k^i)$. We note that $\Dh_k$ can be computed from the $k$th epoch. We would like to bound $\Dh_k - D_k$ as
\[
|\Dh_k - D_k| = \frac{1}{T_k}\sum_{i=1}^T \ell_k^i(\w_k^i) - \ell(\w_k^i)
\]
Using the bound for $A_T$, we have, with a probability $1 - T^2$
\[
|\Dh_k - D_k| \leq 2\sqrt{\frac{Ct}{T_k} D_k} + \frac{\sqrt{2}}{3} \frac{Ct}{T_k}
\]
or
\[
D_k \leq \frac{C}{T^3}
\]
In the second case, since $\E[\Dh_k] = D_k \leq C/T^3$, using the Markov inequality, we have, with a probability $1 - T^{-2}$,
\[
|\Dh_k - D_k| \leq \frac{C}{T}
\]
Combining the above two statements, we have, with a probability $1 - 2T^{-2}$
\[
|\Dh_k - D_k| \leq 2\sqrt{\frac{Ct}{T_k}D_k} + 2\frac{Ct}{T_k}
\]
and consequentially,
\[
|\Dh_k - D_k| \leq 6\left(\sqrt{\frac{Ct}{T_k}\Dh_k} + \frac{Ct}{T_k}\right)
\]
We thus will use the following expression as the surrogate for $\ell(\w_*)$
\[
\lh_k = \Dh_k + 6\left(\sqrt{\frac{Ct}{T_k}\Dh_k} + \frac{Ct}{T_k}\right)
\]
Using $\lh_k$, we define the step size $\eta_{k+1}$ as
\[
\eta_{k+1} = \frac{R}{2\sqrt{\gamma T_{k+1}\lh_k}}
\]
It is easy to verify that with a probability $1 - 2T^{-2}$ (i) $\lh_k \geq D_k \geq \ell(\w_*)$ and (ii) $\lh_k - \ell(\w_*) \leq (K+6) \left(\sqrt{\frac{Ct}{T_k}\Dh_k} + \frac{Ct}{T_k}\right)$. Using the bound in (\ref{eqn:bound-1}), we have
\[
(1 - 2\eta\gamma)(D_{k+1} - \ell(\w_*)) \leq \frac{R^2}{2\eta_{k+1} T_{k+1}} + 2\eta_{k+1}\gamma \ell(\w_*) + 2\sqrt{\frac{Ct}{T_{k+1}}\ell(\w_*)} + 2\sqrt{\frac{Ct}{T_{k+1}}(D_{k+1} - \ell(\w_*))} + \frac{Ct}{T_{k+1}}
\]
Using the property $\lh_k \geq \ell(\w_*)$, we have
\[
2\eta_{k+1} \gamma \ell(\w_*) \leq R\sqrt{\frac{\gamma}{T_{k+1}} \ell(\w_*)}
\]
We also have
\[
\frac{R^2}{2\eta_{k+1}T_{k+1}} = R\sqrt{\frac{\gamma}{T_{k+1}}\lh_k} \leq R\sqrt{\frac{\gamma}{T_{k+1}}(K + 6)\left[\frac{Ct}{T_k} + \sqrt{\frac{Ct}{T_k}\Dh_k} \right]}
\]
Since
\[
\Dh_k - D_k \leq 2\sqrt{\frac{Ct}{T_k} D_k} + 2\frac{Ct}{T_k}
\]
and
\[
D_k - \ell(\w_*) \leq K\left(\frac{Ct}{T_k} + \sqrt{\frac{Ct}{T_k}\ell(\w_*)}\right)
\]
we have
\[
\Dh_k \leq \ell(\w_*) + K\left(\frac{Ct}{T_k} + \sqrt{\frac{Ct}{T_k}\ell(\w_*)}\right) + 2\frac{Ct}{T_k} + 2\sqrt{K}\frac{Ct}{T_k}  + 4\sqrt{K}\frac{Ct}{T_k} + 4\sqrt{K}\sqrt{\frac{Ct}{T_k}\ell(\w_*)}
\]
By choosing sufficiently large $K$, we have
\[
\Dh_k \leq \ell(\w_*) + 2K\left(\frac{Ct}{T_k} + \sqrt{\frac{Ct}{T_k}\ell(\w_*)}\right)
\]
Hence,
\[
\frac{R^2}{2\eta_{k+1}T_{k+1}} \leq \frac{R\sqrt{2\gamma (K+6)Ct}}{T_{k+1}} + \frac{2R^2\gamma}{T_{k+1}} + 2\sqrt{\frac{2Ct}{T_{k+1}}\ell(\w_*)} + 6\sqrt{2K}\frac{2Ct}{T_{k+1}} + 4\sqrt{2K}\sqrt{\frac{2Ct}{T_{k+1}\ell(\w_*)}}
\]
By choosing sufficiently large $K$, we have
\[
\frac{R^2}{2\eta_{k+1}T_{k+1}} \leq \frac{K}{3}\left(\frac{Ct}{T_{k+1}} + \sqrt{\frac{Ct}{T_{k+1}}\ell(\w_*)} \right)
\]
We thus have
\[
(1 - 2\eta\gamma)(D_{k+1} - \ell(\w_*)) \leq \frac{2K}{3}\left(\frac{Ct}{T_{k+1}} + \sqrt{\frac{Ct}{T_{k+1}}\ell(\w_*)} \right)
\]
By choosing $\eta \leq 1/[6\gamma]$, we have
\[
D_{k+1} - \ell(\w_*) \leq K\left(\frac{Ct}{T_{k+1}} + \sqrt{\frac{Ct}{T_{k+1}}\ell(\w_*)} \right)
\]
\bibliography{minimize-binary-loss}
\end{document}